\documentclass[conference]{IEEEtran}
\usepackage{cite}
\usepackage{amsmath,amssymb,amsfonts}
\usepackage{algorithmic}
\usepackage{graphicx}
\usepackage{textcomp}
\usepackage{multirow}
\usepackage{url}
\usepackage{comment}

\begin{document}
\title{Advanced Deep Regression Models for Forecasting Time Series Oil Production}
\author{Siavash Hosseini, Thangarajah Akilan }

\maketitle

\begin{abstract}
Global oil demand is rapidly increasing and is expected to reach 106.3 million barrels per day by 2040. Thus, it is vital for hydrocarbon extraction industries to forecast their production to optimize their operations and avoid losses. Big companies have realized that exploiting the power of deep learning (DL) and the massive amount of data from various oil wells for this purpose can save a lot of operational costs and reduce unwanted environmental impacts.
In this direction, researchers have proposed models using conventional machine learning (ML) techniques for oil production forecasting. However, these techniques are inappropriate for this problem as they can not capture historical patterns found in time series data, resulting in inaccurate predictions. This research aims to overcome these issues by developing advanced data-driven regression models using sequential convolutions and long short-term memory (LSTM) units. 
Exhaustive analyses are conducted to select the optimal sequence length, model hyperparameters, and cross-well dataset formation to build highly generalized robust models.  A comprehensive experimental study on Volve oilfield data validates the proposed models. It reveals that the LSTM-based sequence learning model can predict oil production better than the 1-D convolutional neural network (CNN) with mean absolute error (MAE) and $R^{2}$ score of 111.16 and 0.98, respectively. It is also found that the LSTM-based model performs better than all the existing state-of-the-art solutions and achieves a $37\%$ improvement compared to a standard linear regression, which is considered the baseline model in this work.


\end{abstract}

\begin{IEEEkeywords}
1-D CNN, Volve oilfield, LSTM, Deep learning, time series forecasting.
\end{IEEEkeywords}

\section{Introduction}
\label{sec:introduction}

\IEEEPARstart{T}{he} 18th century marked the first profound industrial revolution that predominantly exploited steam power replacing animal labor.
Since then, there has been rapid development in industrial operations~\cite{6714496}.
Now, the world has come to the brink of the fifth industrial revolution, a.k.a. industry 5.0, where smart systems are built to perform complex tasks more efficiently by leveraging advanced technologies, such as big data, high-performance computing (HPC) platforms, and data-driven analytics~\cite{9695219,9122412}. Thus, industries are increasingly striving to create new and efficient methods of production by utilizing the capabilities of artificial intelligence (AI). These advanced technologies offer a wide range of potential benefits, including increased automation, improved decision-making, and enhanced ability to process and analyze large amounts of data. Hence, the DNNs have become a cornerstone of several industrial operations, including accurate prediction or concept classification of operational conditions, aiming at smart control, real-time fault detection, and maintenance.
For instance, in the oil and gas industry, intelligent assistive tools (IATs) for production forecasting based on readily accessible parameters is crucial for economic assessment and gain.
Nevertheless, it is a challenging task due to (i) the complexity of the environmental and geographical subsurface conditions, (ii) the non-linear relationship between production volume and petro-physical parameters, such as permeability and density, and (iii) the shortage of curated data availability. 
Therefore, despite technological advancement, hydrocarbon production analysis remains an active research field. It urges the research community to develop reliable and precise predictive models. Such models should provide more comprehension of the ongoing production, resulting in efficient operation, and informed decision-making and management.
This work pragmatically develops two deep-learning models using 1-D CNN and LSTM to forecasting oil production. The main contributions of this work are summarized as follows.
\begin{itemize}
    \item Comprehensive study of data pre-processing, viz. handling missing values, data scaling, and feature selection based on petrochemical industrial expertise.
    \item Systematic analysis of time series data for optimal sequence generation.
    \item Hyper-parameter optimization by investigating the most effective model parameters.
    \item Generalized model development.
    \item Exhaustive ablation study and comparative analysis to validate the proposed models' performances.
\end{itemize}

It is worth mentioning that these highlighted contributions are fully or partially missing in the existing works conducted  by other researchers on the same data collected from Volve oil field.  
Thus, this study aims to bridge the main research gaps and propose new strategies to improve production forecasting in the hydrocarbon industries.
The rest of this paper is organized as follows. Section~\ref{sec-lieterature} reviews important relevant works, Section~\ref{sec-method} elaborates on the proposed models, 
and Section~\ref{sec-experiments} presents the methodology, 
Finally, Section~\ref{sec-Overall analysis} and Section~\ref{sec-conclusion} provide an overall summary and conclusion of the paper with future directions, respectively.

\section{Related Works}\label{sec-lieterature}

\begin{figure*}
 \centering
 \includegraphics[width=1\textwidth]{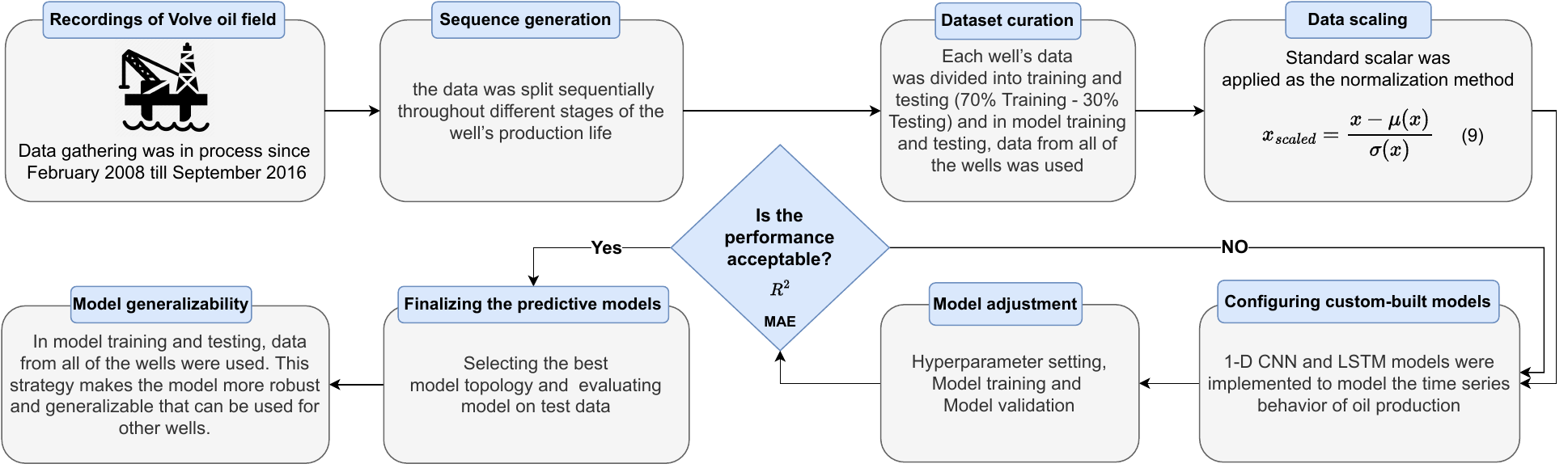}
 \caption{An illustration of the phases involved in building the advanced deep learning-based oil production forecasting models. It subsumes several operations, including data gathering, data curation, model configuration, model training, and model evaluation.} 
 \label{figure:flow_chart}
\end{figure*}

This section overviews the existing related works under two categories: the general application of ML models in hydrocarbon industries for purposes other than production forecasting - Section~\ref{sec:ml-for-general} and the ML models developed exclusively for production forecasting - Section~\ref{sec:forcasting}.

\subsection{Adaptation of ML In Hydrocarbon Industries}\label{sec:ml-for-general}

In the modern era, industries want to explore more pathways for cost saving, increasing productivity, and enhancing safety in the operational environment. Thus, engineers and scientists develop various IATs to facilitate industries in achieving their desired goals. 
For example, in recent years, for solving several problems related to the oil and gas industries, researchers have explored a combination of the nature-inspired meta-heuristic algorithm (MA) and ML techniques. These algorithms are found to have robust performances and converge to the global optimum solution~\cite{bahiraei2021neural, bahiraei2021predicting, ezugwu2020conceptual}. 

On the other hand, Alakeely~\textit{et al.} implemented a recurrent neural network (RNN)-based model along with convolutional neural networks (CNNs) to simulate reservoir behavior~\cite{alakeely2020simulating}. 
In addition to reservoir engineering, some research works focused on applying ML for drilling and construction engineering in the petrochemical industry. For instance, Syed~\textit{et al.}~\cite{syed2020artificial} investigated ML models to predict lift selection, assess the wells’ performance, and to classify them as "Good" or "Bad" wells based on their life-cycle cost (LCC).
Similarly, Adedigba~\textit{et al.}~\cite{adedigba2018data} conducted research touching upon risk assessment of drilling operations using a Bayesian tree augmented Na\"{\i}ve Bayes (TAN). They developed this model to predict time-dependent blowout risk based on the current status of the key drilling parameters in real-time. Hence, it is intended for informed decision-making to avoid preventable workplace accidents and enhance the safety of drilling operations. Furthermore, Ozbayoglu~\textit{et al.} proposed an artificial neural network (ANN)-based model to estimate flow rate and velocity of pipe rotation for real-time drilling optimization and automation~\cite{ozbayoglu2021optimization}. 

\subsection{ML for Forecasting Hydrocarbon Production}\label{sec:forcasting}

ML-driven data analytics-- a branch of science taking advantage of advanced statistical and neural network techniques-- is used to realize and unearth insights and trends in large-scale datasets. It can be potentially exploited to drive meaningful information from hydrocarbon well's raw data aiming at increasing production efficiency and maximizing the profit of petrochemical industries. 
For instance, Bao~\textit{et al.}~\cite{bao2020data} investigated the performance of RNN combined with an ensemble Kalman filter (EnKf) for predicting production to assist reservoir characterization and development. They verified their model on synthetic historical production data, rather than real data collected from an oil field. 
On the contrary, some researchers attempted in developing production forecasting models using actual data. For example, Zanjani~\textit{et al.}~\cite{zanjani2020data} developed multiple algorithms based on ANNs, support vector machine (SVM), and linear regression (LR) for production forecasting using well-specific information. Their results on well NO 159 F-1 C in the Volve oil field show that the ANN-based model performs better than the other two algorithms. Since they focus on single well-specific model development, their approach is not scalable. 




Wang~\textit{et al.}~\cite{wang2021ensemble} conducted a study using machine learning to predict future production. In this research, a machine learning algorithm called the random forest ensemble was implemented to predict time-lapse oil saturation profiles. The algorithm was optimized using feature selection based on feature importance scores and Pearson correlation coefficients in combination with geophysical domain knowledge. The workflow was demonstrated using data from a structurally complex, heterogeneous, and heavily faulted offshore reservoir and was able to predict future time-lapse oil saturation profiles with high accuracy, as measured by over 90\% R-square. This approach is notable because it does not require input parameters derived from cores, petrophysical logs, or seismic data and incorporates production data, which is an essential reflection of dynamic reservoir properties and is typically the most frequently and reliably measured quantity throughout the life of a field.

Li~\textit{et al.} conducted a study for pressure prediction by combining the physics of well's behavior and deep learning (DL) models. Gated Recurrent Unit (GRU) and LSTM models were implemented to compare their results with the RNN model. Results showed that GRU and LSTM performed better compared to RNN. Well NO 15/9-F-1 C from April 2014 to April 2016 was used as a testing profile \cite{li2019deep}.
Masina \textit{et al.} studied automated declined curve analysis (DCA) using AI to predict production rate. Their results showed that the DCA method is able to predict the desired output with a goodness of fit of 0.82 on the test set \cite{masini2019decline}.

Zhang~\textit{et al.}~\cite{8263170} proposed a method for detecting and locating leaks in liquid pipelines, which combines inverse hydraulic-thermodynamic transient analysis with an improved version of the particle swarm optimization (PSO) algorithm. The finite volume method is used to solve the continuity, momentum, and energy equations numerically. Four different algorithms were tested to determine the best-performing version of the improved PSO algorithm, and the results were evaluated based on accuracy, stability, robustness, and false alarm rate. The SIPSO algorithm was found to be the most effective. The proposed method was applied to two oil pipelines in real-world scenarios, one during a field opening experiment and the other during a leak incident. The method was able to accurately estimate the location, coefficient, and starting time of the leaks with low relative errors.

Noshi \textit{et al.} explored the potential application of Machine Learning algorithms
in production prediction. They took advantage of the AdaBoost technique for production prediction. Mean absolute error (MAE) was used as an error metric to show the method's performance. Six features that affect production prediction, including: on stream hours, average choke size, bore oil volume, bore gas volume, bore water volume, and finally, average wellhead pressure were used as input parameters \cite{noshi2019intelligent}.

Panja \textit{et al.} carried out a study to predict hydrocarbon
production from hydraulically fractured wells. Two common types of ML models, namely the Least Square Support Vector Machine (LSSVM) and the Artificial Neural Networks (ANN) were analyzed and compared to the traditional curve fitting method known as Response Surface Model (RSM) using second-order polynomial equations to determine production rate \cite{panja2018application}.
Wui Ng \textit{et al.} studied LSTM model for Volve oilfield production forecasting. In the mentioned work, only well NO 15/9-F-14 H were used for both the training and testing process \cite{ng2022well}. This paper used an incorrect strategy in methodology. Specifically, the correlation between the production of oil and gas was found to be equal to 1, but the authors used gas as the input in their network and oil as the output. A more appropriate approach would have been to remove gas from the input variables in order to avoid this issue as it is highly correlated with the output and will result in a bias in the predictions made by the network.

One of the classical methods which was used for hydrocarbon production forecasting in recent decades is decline curve analysis (DCA). This method was initiated by Arps~\textit{et al.} (1945), and then oil and gas companies adopted this method and its specific applications in related industries \cite{arps1945analysis}. As a result of its simple development, it has been broadly used in various situations \cite{hong2019integrating}. For forecasting hydrocarbon production, numerical reservoir simulation (NRS) can be used as an alternative to DCA. Yet, the performance of the NRS method depends on how historical matching (HM) has been done \cite{liu2019petroleum}. Also, NRS requires several features, comprising well locations, fluid properties, geological data, etc. To forecast production more accurately, the simulation model should be updated via HM when new real-time data emerges. As a result, this method has apparent limitations \cite{ng2022well}.

Data-driven modeling has become a viable option for hydrocarbon production forecasting with the advancement of data analytics and computing technology. Not only is this method easy to implement, but it also captures the intricate relationships between inputs and outputs. Utilizing machine learning (ML) has led to notable advancement in the oil and gas industry, especially in reservoir engineering \cite{ng2022well}.
However, the hydrocarbon industries face challenges in accomplishing this, as the existing conventional tools are not generalized and robust enough. In recent decades, AI and DL-based innovative solutions have emerged to improve the efficiency of operations in industries.
This research has been carried out, which aims to propose two models based on convolutional neural network (CNN) and long short-term memory (LSTM) for hydrocarbon production. 
The following section (Methodology) elaborates on all of the stages that contribute to the production forecasting of the Volve oilfield data set.

\begin{figure*}[!htp]
 \centering
 \includegraphics[width=1\textwidth]{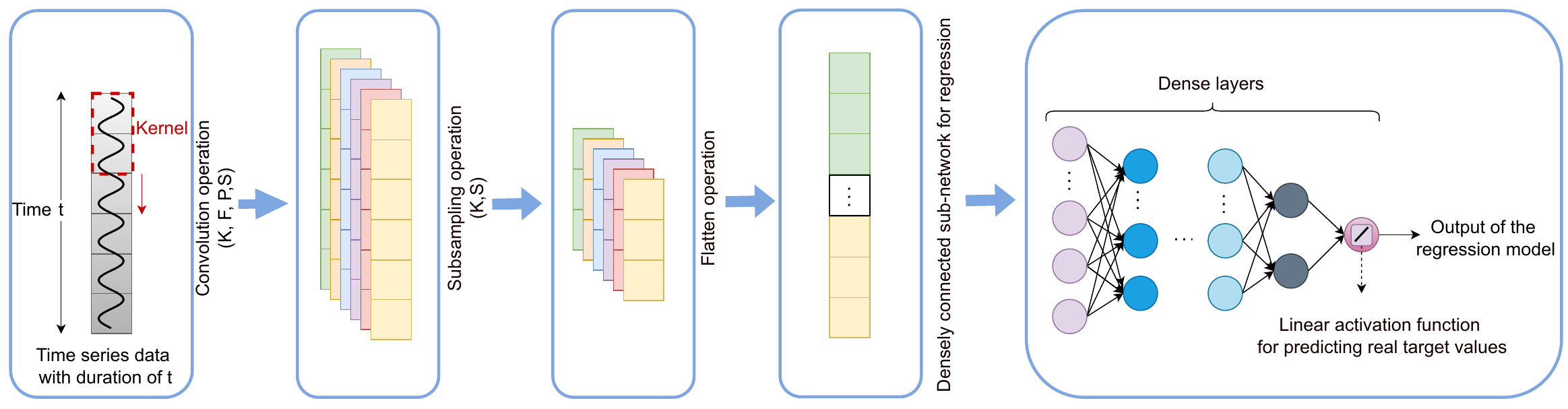}
 \caption{An illustration of sequential regression model using 1-D CNN. It subsumes input sequence learning using 1-D Conv, subsampling using max pooling, and regression output generation via a densely connected subnetwork, where K, F, P, and S stand for conv kernel size, the number of filters, padding, and stride rate, respectively.}
 \label{figure:1dcnn}
  \vspace{-0.5cm}
\end{figure*}

\section{Methodology}\label{sec-method}


In the last few decades, there has been a significant amount of focus on improving the architecture of DNNs. This attention is due to the fact that DNNs have been effective in solving a wide range of practical problems that arise in various industries. 
One of the most significant advantages of using these networks is their high capability to learn non-linear relationships regardless of the type of data~\cite{hosseini2020precise, 9349558,bahiraei2021neural,hosseini2021increasing, li2022global, hosseini2022accurate, 7946733, hosseini2021application}.
As a result, there is a growing interest among researchers to optimize the structure of DNNs. 
This has involved the exploitation of different topologies, such as skip connections, the application of various techniques to reduce the number of trainable parameters, and novel fast retraining to fine-tune pre-trained DNNs~\cite{8718406}. 

In this direction, this work advances the regression models for hydrocarbon production forecasting by exploiting the power of the learning capability of the DNNs, particularly in time-series data analysis. 
This work adopts a modeling technique where it progressively designs and develops predictive models from a baseline least squares-based linear regression to advanced deep learning-based regression models. Fig.~\ref{figure:flow_chart} illustrates the phases involved in the building of the advanced deep learning-based solutions. The following subsections elaborate on the proposed solutions in a step-by-step manner.

\subsection{Base Model: A Standard Linear Regressor}\label{sec:baseline} 
Linear regression (LR) estimates the linear relationship between different explanatory attributes and a dependant variable of given data samples (cf.~\ref{eq:linear}) by minimizing an objective function, say the sum of the squares, i.e., the distance between each predicted and actual value of the dependent variable is squared and then summed up for all training samples. 
Due to its well-established mathematical foundation and easy training procedure, LR is widely accepted as a baseline model for various regression problems in several fields, including engineering, biomedical, behavioral, and social sciences, and business. The standard linear regression model can be defined as in \eqref{eq:linear}.
\begin{equation}
    \label{eq:linear}
    y = \alpha_{1} x_{1}+\alpha_{2} x_{2}+...+\alpha_{n} x_{n}+\beta,
\end{equation}
where $y$, $\alpha_{i}$, $x_{i}$, and $\beta$ stand for the dependent variable (output), coefficient of the $i$th input attribute, $i$th input attribute, and bias, respectively.
The coefficients are optimized by minimizing the total sum of squares (SST) defined in \eqref{eq:sst}, which is the aggregation of the sum of squares (SSE), $\sum_{i=1}^{n}(y_{i}-\widehat{y_{i}})^{2}$ and the sum of squares (SSR), $\sum_{i=1}^{n}(\widehat{y_{i}}-\overline{y})^{2}$. 
\begin{equation} \label{eq:sst}
\begin{split}
 \sum_{i=1}^{n}(y_{i}-\overline{y})^{2} & =\sum_{i=1}^{n}(y_{i}-\widehat{y_{i}})^{2} + \sum_{i=1}^{n}(\widehat{y_{i}}-\overline{y})^{2},
\end{split}
\end{equation}
where ${y_{i}}$ stands for the real value in observation ${i}$, ${\overline{y}}$ is the mean value of the dependant variable, $y$ from all $n$ observations, and ${\widehat{y_{i}}}$ denotes the predicted value of the dependant variable for the given $i$th observation's input attributes.

\begin{table}[!t]
\caption{Architecture Detail of the 1-D CNN-based Regressor}
\label{table:cnn_char}
\centering
\begin{tabular}{ccc}
\hline
\multicolumn{1}{c|}{Layer ID} & \multicolumn{1}{c|}{Layer Type}  & \multicolumn{1}{c}{Output Dimension}  \\ \hline
\multicolumn{1}{c|}{Input}    & \multicolumn{1}{c|}{Input layer} & \multicolumn{1}{c}{(32, 5, 12)}   \\
\multicolumn{1}{c|}{L1}       & \multicolumn{1}{c|}{Conv 1D}  & \multicolumn{1}{c}{(32, 4, 64)}   \\
\multicolumn{1}{c|}{L2}       & \multicolumn{1}{c|}{Conv 1D} & \multicolumn{1}{c}{(32, 3, 64)}    \\
\multicolumn{1}{c|}{L3}       & \multicolumn{1}{c|}{Conv 1D} & \multicolumn{1}{c}{(32, 2, 64)}    \\
\multicolumn{1}{c|}{L4}       & \multicolumn{1}{c|}{MaxPooling 1D} & \multicolumn{1}{c}{(32, 1, 64)}    \\
\multicolumn{1}{c|}{L5}       & \multicolumn{1}{c|}{Flatten}  & \multicolumn{1}{c}{(32, 64)}       \\
\multicolumn{1}{c|}{L6}       & \multicolumn{1}{c|}{Dense} & \multicolumn{1}{c}{(32, 1)}        \\ \hline
\multicolumn{3}{l}{\begin{tabular}[c]{@{}l@{}} Total number of trainable parameters: 18,049\hspace{3cm}\hfill \\ 
Number of filters (F) in L1, L2, and L3: 64\\ Kernel size (K) in L1, L2, L3, and L4: 2\\Activation function: L1-L5: ReLu; L6: linear\\ Stride rate (S) was equal to 1 in all layers\\ Learning rate: 0.001; Number of epochs: 200\\
Optimizer: Adam; Batch size: 32
\end{tabular}} \\ \hline
\end{tabular}
\vspace{-0.2cm}
\end{table}

\subsection{1-D CNN-based Regressor} 


CNNs have been known for their robustness and become the de facto standard in a wide range of computer vision tasks~\cite{9050859, 8718406, 8671459}. 
One unique feature that makes CNNs efficient for supervised learning is the spatial-local connectivity that allows layers to share parameters \cite{9918222}. Feature extraction in CNNs relies heavily on the convolution (Conv) layers, which perform convolution operations on the input data or input feature map(s) using pre-configured kernels, i.e., feature detectors as defined in~\eqref{eq:1dcnn}. This Conv operation will generate a volume of learned feature maps.
In this work since the input is 1-D sequential data, the input Conv layer receives a 1-D input sequence, $x(n) \in \mathbb{R}^{1\times6}$. Then, then a convolution between the kernel, ${w(n)}$ and the input generates a feature map, $z(n)$ as defined in \eqref{eq:1dcnn}~\cite{zhao2019speech, kiranyaz20191}.
\begin{equation}
    \label{eq:1dcnn}
    z(n) = x(n)\ast w(n) = \sum_{m=-k}^{k}x(m)\cdot w(n-m),
\end{equation}
where ${k}$ and '\textasteriskcentered' denote the kernel size and Conv operation, respectively.
The hyper-parameters, such as the number of hidden layers, kernel size (K), number of filters (F), sub-sampling factor, and type of activation function used in each layer determine the structure of the 1-D CNN model. In this work, the proposed 1-D CNN regressor's structure, its layer connectivity, and the hyperparameter setting are given in Fig.~\ref{figure:1dcnn} and Table~\ref{table:cnn_char}. 

\subsection{LSTM-based Regressor}\label{lstm-regressor}

\begin{table}[!t]
\caption{Architecture Detail of the LSTM-based Regressor}
\label{table:lstm_char}
\centering
\begin{tabular}{ccc}
\hline
\multicolumn{1}{c|}{Layer ID}   & \multicolumn{1}{c|}{Layer Type}    & \multicolumn{1}{c}{Output Dimension}       \\ \hline
\multicolumn{1}{c|}{Input}      & \multicolumn{1}{c|}{Input layer}   & \multicolumn{1}{c}{(32, 5, 12)}  \\
\multicolumn{1}{c|}{L1}         & \multicolumn{1}{c|}{LSTM}          & \multicolumn{1}{c}{(32, 5, 30)}   \\
\multicolumn{1}{c|}{L2}         & \multicolumn{1}{c|}{LSTM}          & \multicolumn{1}{c}{(32, 20)}   \\
\multicolumn{1}{c|}{L3}         & \multicolumn{1}{c|}{Dense}         & \multicolumn{1}{c}{(32, 20)}   \\
\multicolumn{1}{c|}{L4}         & \multicolumn{1}{c|}{Dense}         & \multicolumn{1}{c}{(32, 16)}   \\
\multicolumn{1}{c|}{L5}         & \multicolumn{1}{c|}{Dense}         & \multicolumn{1}{c}{(1)}  \\ \hline
\multicolumn{3}{l}{\begin{tabular}[c]{@{}l@{}}Total number of trainable parameters: 9,893\hspace{3cm}\hfill\\ 
Number of hidden units in L1  and L2: 30 and 20\\Activation function: L1, L2: tanh; L3, L4: Relu; L5: linear\\ Learning rate: 0.001;
Number of epochs: 200\\
Optimizer: Adam;
Batch size: 32
\end{tabular}} \\ \hline
\end{tabular}
 \vspace{-0.5cm}
\end{table}

\begin{figure}[t]	
	\centering
\includegraphics[trim={3.5cm, 21.5cm, 9.7cm, 2cm}, clip, width=1\columnwidth]{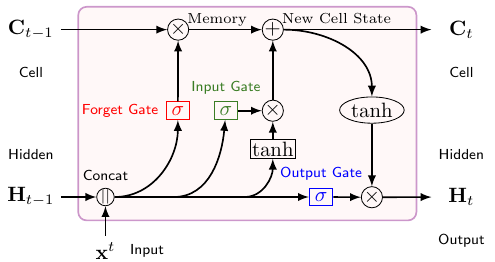}

$\bigoplus$ - pointwise addition, $\bigotimes$ - pointwise multiplication
	\caption{An illustration of a standard LSTM cell with three gates that control information flow, where $\mathbf{X}_t$, $\mathbf{C}_t$, and $\mathbf{H}_t$, are the input quantity from a time-series data, cell state, and hidden state, respectively, at timestamp $t$.}
  	\vspace{-0.2cm}
	\label{fig:lstm_cell}
\end{figure}

LSTM networks are an improved version of recurrent neural networks (RNNs) mainly introduced to better handle the long short-term dependencies sequential data. 
LSTM-based models have proven to be the state-of-the-art in several time series analysis, viz. stock market prediction~\cite{chen2015lstm}, moving object detection~\cite{8671459} and speech recognition~\cite{graves2005framewise}. The recurrent connections and memory mechanisms play a vital role in LSTMs to retain significant past observations. By taking advantage of the input gate, forget gate, and output gate operations as defined in \eqref{eqn_lstm-input} - \eqref{eqn_lstm-hidden}, it is possible for an LSTM module shown in Fig.~\ref{fig:lstm_cell} to add new useful information to the memory and omit some information that is no longer important to be preserved. 
\begin{gather} \label{eqn_lstm-input}
i_t = \sigma(W_{xi} \ast X_t + W_{hi} \ast H_{t-1} + b_i),\\
f_t = \sigma(W_{xf} \ast X_t + W_{hf} \ast H_{t-1} +b_f),\\
o_t = \sigma(W_{xo} \ast X_t + W_{ho} \ast H_{t-1} +b_o),\\
C_t = f_t \circ C_{t-1} + i_t \circ \tanh(W_{xc} \ast X_t + W_{hc} \ast H_{t-1} +b_c),\\
H_t = o_t \circ \tanh(C_{t}),\label{eqn_lstm-hidden}
\end{gather}
where $X_t$ is an input from a time-series data, $C_t$ is the cell state, $H_t$ is the hidden state, and $i_t$, $f_t$, and $o_t$ are the gates of the LSTM module at timestamp $t$. Hence, $W$, '\textasteriskcentered', '$\circ$', and $\sigma$ denote the Conv kernels specific to the gates or the internal states, the Conv operator, the Hadamard product, i.e., element-wise matrix multiplication, and \texttt{hard sigmoid} activation function.


\vspace{0.5cm}

\begin{figure}[!t]	
	\centering
\includegraphics[width=1\columnwidth]{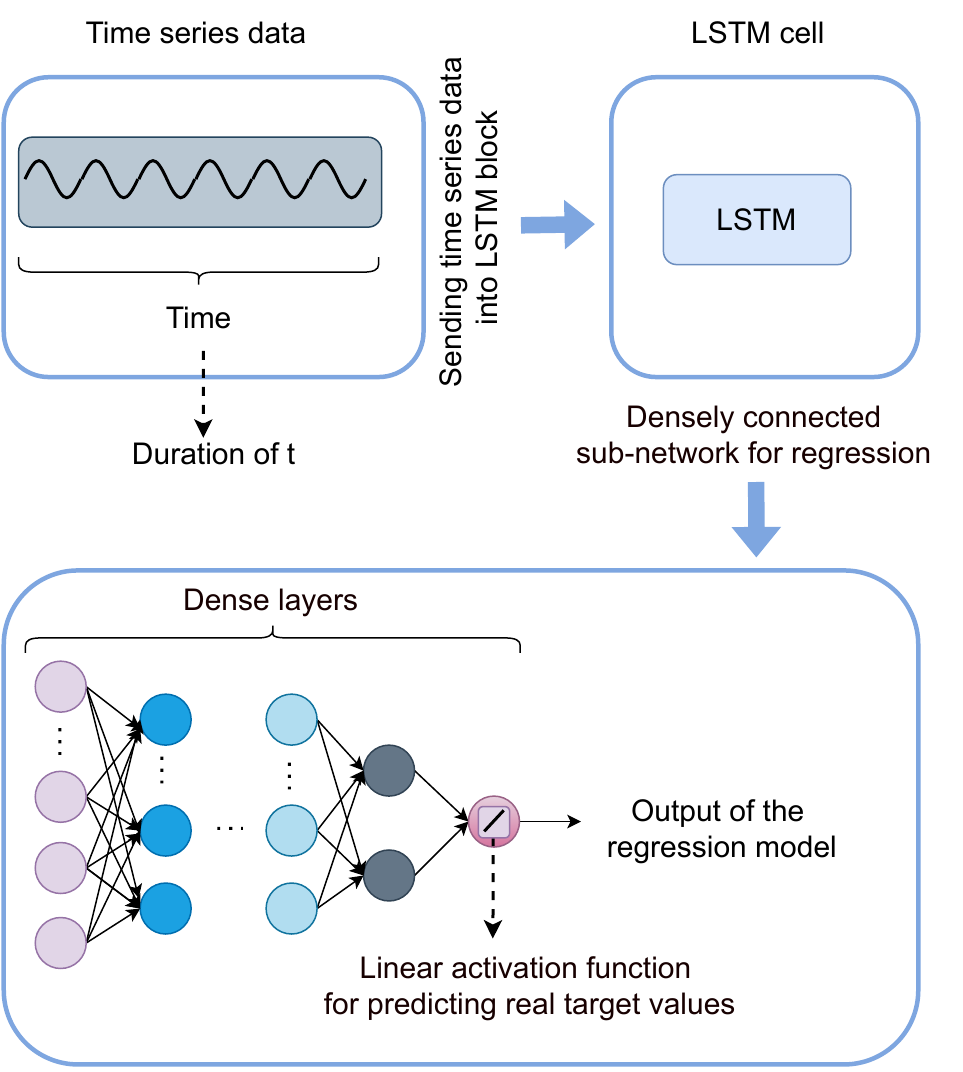}
	\caption{An overview of applying LSTM for sequence learning. Here, a time series data with a sequence length of $t$ is input to the LSTM subnetwork. The learnt representation by the LSTM subnetwork is then forwarded to the densely connected regression sub-network that generates the final output.}
	\label{fig:lstmdia}
\end{figure}

While Fig.~\ref{fig:lstmdia} illustrates the general idea of how LSTMs can be applied for a sequence-based regression problem, Table~\ref{table:lstm_char} provides the layer connectivity details of the proposed LSTM-based regressor.

\section{Experimental analysis}\label{sec-experiments}

\subsection{Environment}

This work exploits Google Colaboratory cloud resources with one Tesla T4 graphical processing unit, 12 Gigabytes of RAM, and 2 CPU cores for training, and evaluation. The models are built using Python programming language and the pre-built libraries, like Numpy, Pandas, Matplotlib, Seaborn, and the open-source deep learning library TensorFlow library.

\subsection{Explanatory Data Analysis}\label{sec-EDA}

\subsubsection{Data source}
This work uses the Volve oil field database for the experimental study. The Volve oil field is located in the central part of the North Sea at 2750 - 3120 m depth. The field was revealed in 1993, and the drilling process started in May 2007~\cite{ng2022well}. After 8.5 years, the Volve oil field was decommissioned in 2016~\cite{noshi2019intelligent}. 
In May 2018, Equinor released the Volve database publicly for research and development purposes~\cite{equinor2018volve1}. The database includes different categories of data collected from various operations, but this study focuses on real-field production data. The production data subsumes information gathered from seven wells (five producers and two injectors) as summarized in Table~\ref{table:duration}, namely NO 15/9-F-1 C, NO 15/9-F-11H, NO 15/9-F-12 H, NO 159–F–14 H, NO 15/9-F-15 D, NO15/9-F-4 AH, and NO 15/9-F-5 AH, where 15/9-F-4 and 15/9-F-5 are the two injectors. 

\begin{table}[!t]
\centering
\caption{Historical recordings of different wells in the volve oil field. Note that data was recorded daily from each well.}
\label{table:duration}
\scalebox{0.9}{
\begin{tabular}{cccc}
\hline
\textbf{Well code} & \textbf{Type} & \textbf{Start of recording} & \textbf{End of recording} \\ \hline
NO 15/9-F-1 C    & Producer & 07-Apr-14   & 21-Apr-16 \\
NO 15/9-F-11 H   & Producer  & 08-Jul-13  & 17-Sep-16 \\
NO 15/9-F-12 H   & Producer  & 12-Feb-08  & 17-Sep-16 \\
NO 15/9-F-14 H   & Producer  & 12-Feb-08  & 17-Sep-16 \\
NO 15/9-F-15 D  & Producer   & 12-Jan-14  & 17-Sep-16 \\
NO 15/9-F-4 AH  & Injector   & 01-Sep-07  & 01-Dec-16 \\
NO 15/9-F-5 AH  & Injector   & 01-Sep-07  & 18-Sep-16 \\ \hline
\end{tabular}}
 \vspace{-0.5cm}
\end{table}

\subsubsection{Attributes}
\begin{figure*}[!t]
 \centering
 \includegraphics[width=1\textwidth]{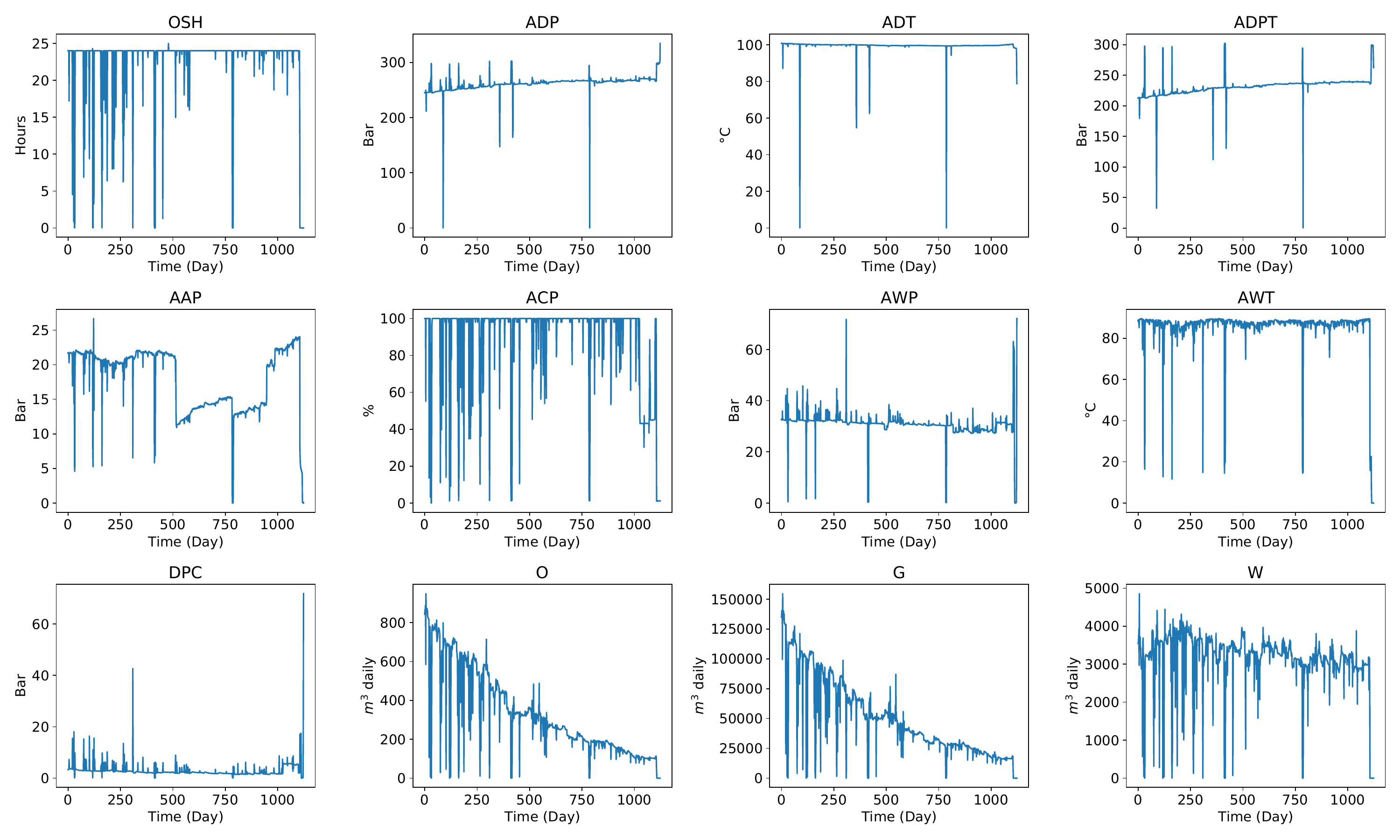}
 \caption{Trend plot visualization of the attributes: OSH, ADP, ADT, ADPT, AAP, ACP, AWP, AWT, DPC, O, G and W for the well no 15/9-F-14 H.}
 \label{figure:well14}
  \vspace{-0.2cm}
\end{figure*}

Table~\ref{table:each_well_data} lists the attributes of the collected data from seven wells stated in Table~\ref{table:duration}. To understand the nature of each attribute, they are visualized using trend plots wrt duration. For example, Fig.~\ref{figure:well14} visualizes the attributes of well no. 14 (NO 15/9-F-14 H). 
Similarly, the attributes' basic statistical information is also analyzed for each well. For instance, Table~\ref{table:statistic} provides the statistical information: mean, standard deviation, minimum value, 1st quartile, median, 3rd quartile, and maximum value of each attribute of well no. 14.



\begin{figure*}[!tp]
 \centering
 \includegraphics[width=1\textwidth]{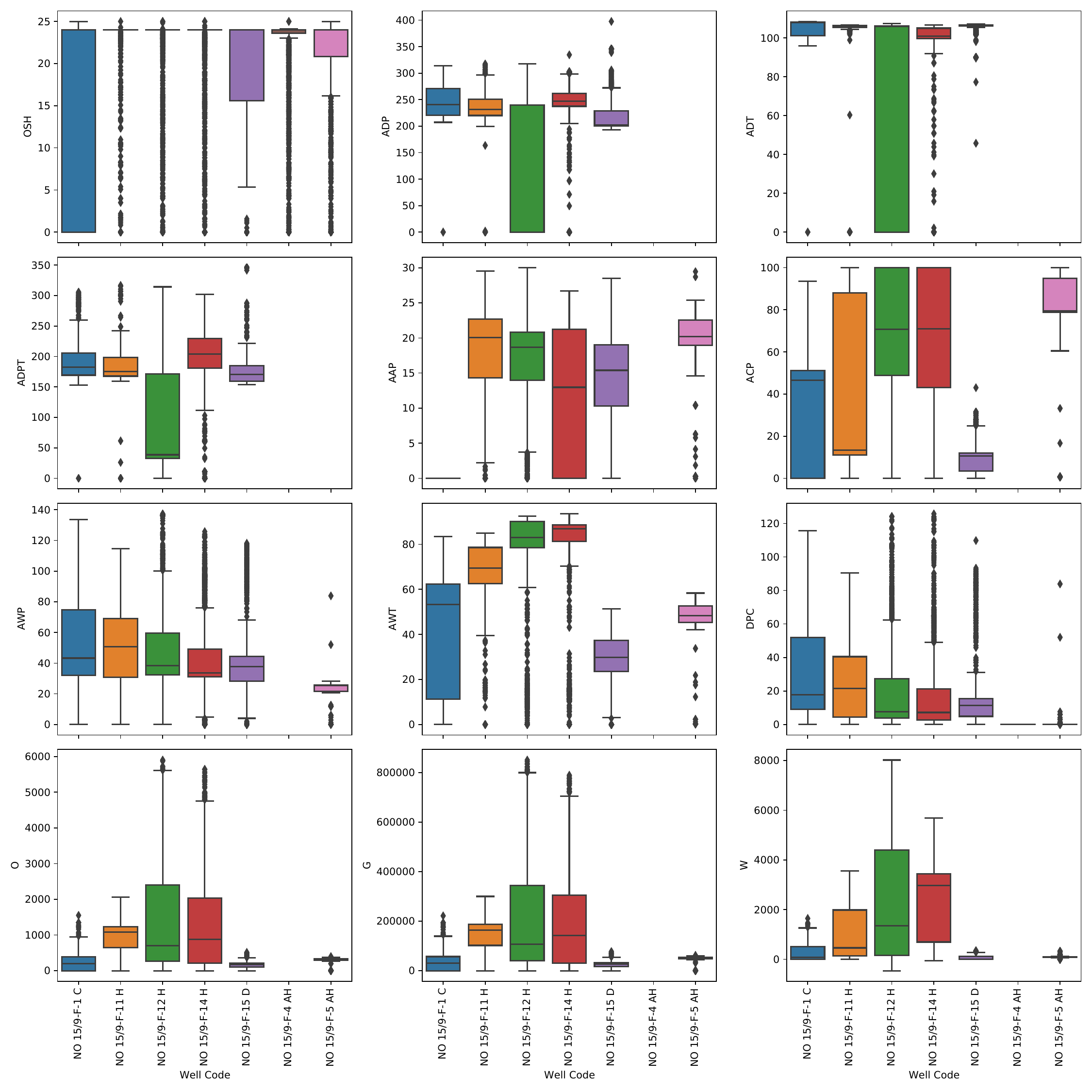}
 \caption{Visualizing the data distribution of the key attributes with respect to each well listed in Table~\ref{table:duration}.}
 \label{figure:boxplot}

\end{figure*}

\begin{table*}[!t]
\centering
\caption{Description of wells' attributes recorded from the volve oil field~\cite{ng2022well}}\label{table:each_well_data}
\begin{tabular}{p{6cm} | p{6cm} | p{2.2cm}}
\hline
\vspace{0.01cm}\textbf{Attributes} & \vspace{0.01cm}\textbf{Description} & \textbf{Abbreviation used in this study} \\ \hline
DATEPRD     & Date of Record        & DOR   \\
ON\_STREAM\_HRS   & On stream hours & OSH   \\
AVG\_DOWNHOLE\_PRESSURE  & Average Downhole Pressure & ADP  \\
AVG\_DOWNHOLE\_TEMPERATURE            & Average Downhole Temperature                                                                & ADT                                \\
AVG\_DP\_TUBING                       & Average Differential Pressure of Tubing                                                     & ADPT                               \\
AVG\_ANNULUS\_PRESS                   & Average Annular Pressure                                                                    & AAP                                \\
AVG\_CHOKE\_SIZE\_P                    & Average Choke Size Percentage                                                               & ACP                                \\
AVG\_WHP\_P                           & Average Wellhead Pressure                                                                   & AWP                                \\
AVG\_WHT\_P                           & Average Wellhead Temperature                                                                & AWT                                \\
DP\_CHOKE\_SIZE
    & Differential Pressure of Choke Size
                                &DPC
        \\

BORE\_OIL\_VOL                        & Oil Volume from Well                                                                        & O                                  \\
BORE\_WAT\_VOL                        & Water Volume from Well                                                                      & W                                  \\
BORE\_GAS\_VOL                        & Gas Volume from Well                                                                        & G                                  \\
BORE\_WI\_VOL                         & Water Volume Injected                                                                       & WI                                 \\
FLOW\_KIND                           & Type of Flow (production or injection)                                                      & TOF                                \\
WELL\_TYPE                           & \begin{tabular}[c]{@{}l@{}}Type of Well (oil production or water injection)\end{tabular} & TOW                                \\ \hline
\end{tabular}
 \vspace{-0.5cm}
\end{table*}

\begin{table*}[!t]
\centering
\caption{Fundamental statistical information of different features of well No 15/9-F-14 H from the volve database}
\label{table:statistic}
\begin{tabular}{|c|c|c|c|c|c|c|c|c|c|c|c|}
\hline
      & OSH      & ADP      & ADT      & ADPT     & ACP      & AAP      & AWP      & AWT      & DPC      & G        & O        \\ \hline
Count & 9001     & 8980     & 8980     & 8980     & 8759     & 7730     & 8995     & 8995     & 8995     & 9001     & 9001     \\
Mean  & 20.1969  & 181.8039 & 77.16297 & 154.0288 & 54.91872 & 14.79186 & 45.80569 & 68.14038 & 19.4882  & 163182.8 & 1110.534 \\
STD   & 8.266102 & 109.7124 & 45.65795 & 76.75237 & 36.67992 & 8.411588 & 24.72463 & 27.67885 & 22.62904 & 189092.9 & 1330.366 \\
Min   & 0        & 0        & 0        & 0        & 0        & 0        & 0        & 0        & 0        & 0        & 0        \\
1st quartile   & 24       & 0        & 0        & 83.66536 & 18.59543 & 10.7977  & 31.42122 & 58.96197 & 3.480234 & 29381.58 & 190.4    \\
Median   & 24       & 232.8969 & 103.1867 & 175.5889 & 51.6073  & 16.18796 & 38.3469  & 80.2844  & 9.888581 & 89678.31 & 574.57   \\
3rd quartile   & 24       & 255.4015 & 106.2766 & 204.32   & 99.9372  & 21.26906 & 57.80183 & 88.11501 & 27.39888 & 208411.8 & 1377.51  \\
Max   & 25       & 397.5886 & 108.5022 & 345.9068 & 100      & 30.01983 & 137.311  & 93.50958 & 125.7186 & 851131.5 & 5901.84  \\ \hline
\end{tabular}
\end{table*}



\subsubsection{Handling missing values}

\begin{table*}[ht!]
\centering
\caption{Sanity Analysis for Finalizing an Architecture for the Proposed LSTM-based and 1-D CNN-based Regressors. Note: $\varnothing$ means that model does not include that layer [A supplementary document was provided for reviewers that includes additional information and details about the models that have been implemented and also submitted along with this manuscript].}
\label{table:hyper_tuning}

\begin{tabular}{|l|l|l|l|l|l|l|l|l|l|l|l|}
\hline
Type & Model ID & Layer 1 & Layer 2    & Layer 3    & Layer 4    & Layer 5 &  Layer 6     &   Layer 7  & FLOPs  & MAE & ${R^2}$ \\ \hline
\multirow{5}{*}{LSTM}    & Model 1          & LSTM    & Dense      & Dense      &     $\varnothing$       &   $\varnothing$      &   $\varnothing$    &  $\varnothing$  & 27552   &  136.37   &   0.97      \\
                         & Model 2          & LSTM    & LSTM       & Dense      & Dense      &    $\varnothing$     &  $\varnothing$     &   $\varnothing$  & 30272  &  113.39   &     0.98    \\
                         & Model 3          & LSTM    & LSTM       & Dense      & Dense      & Dense   &   $\varnothing$    &    $\varnothing$ &  33312 &  111.16   &    0.98     \\
                         & Model 4          & LSTM    & LSTM       & LSTM       & Dense      & Dense   & Dense &   $\varnothing$ & 40352   &  114.58   &    0.97     \\
                         & Model 5          & LSTM    & LSTM       & LSTM       & Dense      & Dense   & Dense & Dense & 48288 &  117.40   &    0.97     \\ \hline
\multirow{5}{*}{1-D CNN} & Model 1          & Conv    & MaxPooling & Flatten    & Dense      &    $\varnothing$     &   $\varnothing$    &   $\varnothing$ & 385056  &  199.88   &     0.94    \\
                         & Model 2          & Conv    & Conv       & MaxPooling & Flatten    & Dense   &   $\varnothing$    &   $\varnothing$ & 1955872   &  322.88   &    0.84     \\
                         & Model 3          & Conv    & Conv       & MaxPooling & Flatten    & Dense   & Dense &   $\varnothing$ & 1993696  &  293.49   &    0.85     \\
                         & Model 4          & Conv    & Conv       & Conv       & MaxPooling & Flatten & Dense &    $\varnothing$ & 3008544  &  151.64   &     0.96    \\
                         & Model 5          & Conv    & Conv       & Conv       & MaxPooling & Flatten & Dense & Dense & 3046368 &  161.34   &    0.95     \\ \hline
\end{tabular}
 \vspace{-0.5cm}
\end{table*}


Handling missing values plays a crucial role in data-driven model building. The Volve oil field database contains missing values in certain attributes. For example, one can find from the statistical information of well no. 15/9-F-14 H summarized in Table~\ref{table:statistic} that there are data samples with missing values in attributes ADP, ADT, ADPT, ACP, AAP, AWP, AWT, and DPC.  
To resolve this, it is important to understand the information distribution of attributes across all the samples. In this case, the distribution is observed using boxplots as shown in Fig.~\ref{figure:boxplot}. From these plots, it is clear that the data is highly skewed, so the best approach for missing value imputation is replacing the missing value with the median value of the respective attribute.

\subsubsection{Feature selection}
\begin{figure}[!t]
\centerline{\includegraphics[trim={2.2cm, 1.8cm, 3.1cm, 2.3cm}, clip, width=\linewidth]{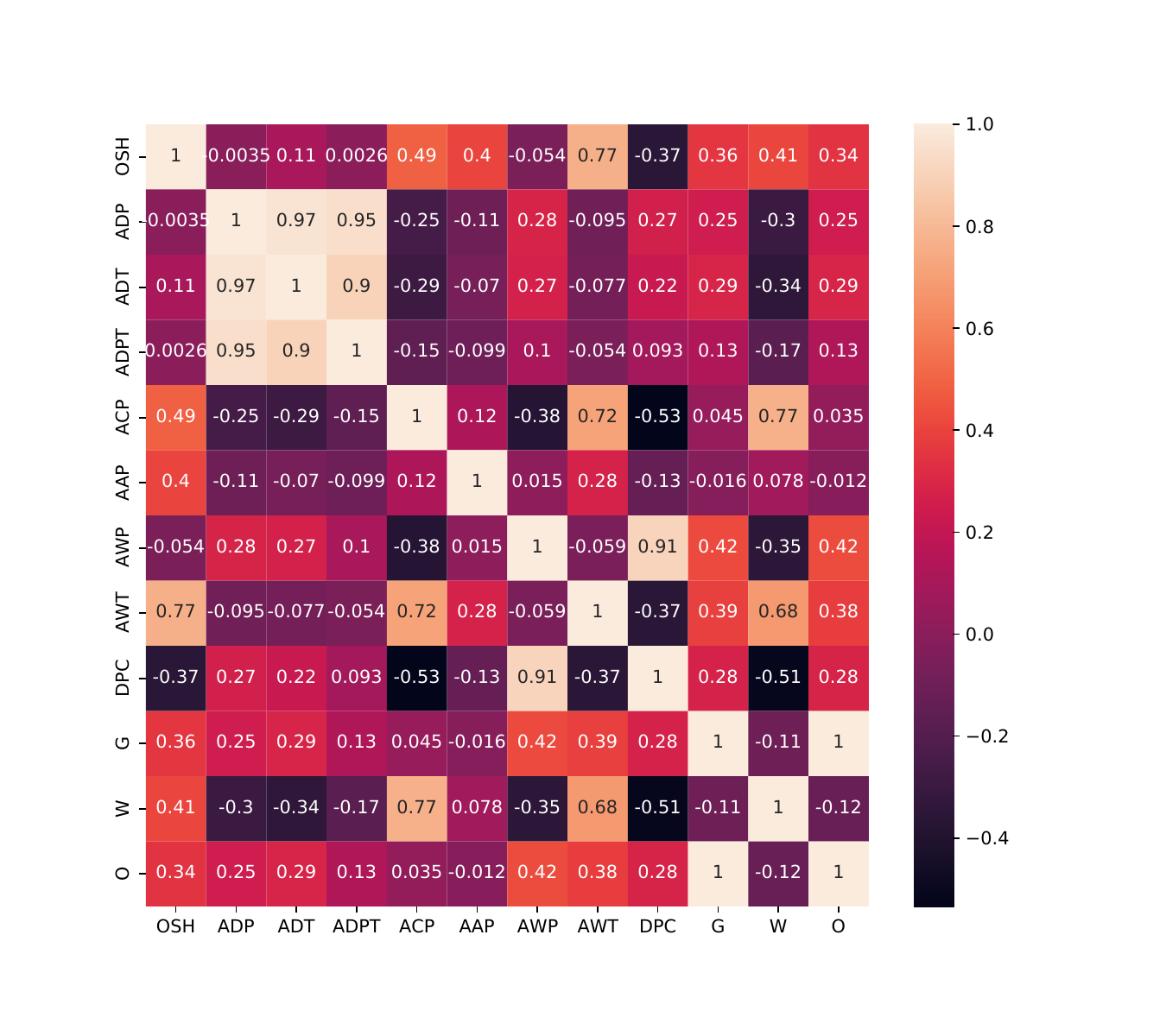}}
\caption{Heatmap of linear correlations between different Features.}
\label{fig:heatmap}
 \vspace{-0.5cm}
\end{figure}

In order to gain a better comprehension of the data, correlations among all of the attributes are calculated and visualized using a heat map shown in Fig.~\ref{fig:heatmap}. 
It is worth mentioning that the correlation between ADP, ADT, and ATPT is equal to 0.97 and 0.95, respectively. 
Therefore, ADP is removed from the input variable list. Moreover, the correlation between produced gas and oil is equal to 1, which complies that oil flow produces gas.



\subsection{Data Preprocessing}

\subsubsection{Data scaling}

Data scaling is a critical preprocessing step, which shows substantial performance gain in time-series analysis~\cite{hosseini2021increasing, a2018effect, bahiraei2020neural}. In this work, standard scalar was applied as the normalization method. This method normalizes the data by subtracting the mean and dividing it by the standard deviation as given in~\eqref{eq:standard_scaler}. It should be noted that  during the process of analyzing the test set, predicted target values are re-scaled to the original range.
\begin{equation}\label{eq:standard_scaler}
{x_{scaled} = \frac{x - \mu(x)}{\sigma(x)}, }
\end{equation}
where $x$, $\mu(x)$, and $\sigma(x)$ are the raw input, the sample mean, and the standard deviation, respectively.

\subsubsection{Dataset curation}

It is found that the existing works on the Volve oil field production database are oil well-specific models; thus, they are not generalized solutions to all the wells. To address this, we curate mutually exclusive training and test datasets that comprise data samples from the five production-related wells listed in Table~\ref{table:duration}. 
In this regard, after generating sequential data samples with a sequence length of $t$, the oil well-specific sequences are divided into $70:30$ non-overlapping training and test datasets. To build generalized models, a global train set and a global test set are formed, respectively by amalgamating all well-specific training sets and test sets. The resulting global sets contain 6286 and 2694 sequential data samples in training and testing sets, respectively. 
This strategy makes the proposed models more accurate, robust, and generalized across all the oil wells compared to the existing solutions. 

\subsection{Evaluation Metrics}

This work uses MAE and R-squared (${R^2}$ score) defined in \eqref{eq:mae} and \eqref{eq:r2} to measure the precision of the oil production forecasts by the proposed models.  
\begin{equation}
    \label{eq:mae}
    MAE=\frac{\sum_{i=1}^{n}\left| y_{i}-\hat{y_{i}}\right|}{n},
\end{equation}
\begin{equation}
    \label{eq:r2}
    R^2 =1-\frac{\sum_{i=1}^{n}(y_{i}-\widehat{y_{i}})^{2}}{\sum_{i=1}^{n}(y_{i}-\overline{y})^{2}},
\end{equation}
where $y_{i}$, $\hat{y_{i}}$ and $\bar{y}$ stand for actual, predicted and average values of target attributes, respectively. Also $n$ denotes to total number of data points.

\subsection{Hyper-parameter Tuning}

Hyperparameter tuning is a meta-optimization task, which plays a vital role in the realm of DL. Effective hyperparameter adjustments have proved to improve predictive models' performances~\cite{JMLR:v13:bergstra12a}. In this case, two important aspects, such as the input sequence length and the layer configuration of the proposed model are considered for hyper-parameter tuning. 

\subsubsection{Optimizing sequence length} 
\begin{figure}[!ht]
 \centering
 \includegraphics[trim={0.5cm, 0.1cm, 1.5cm, 1.1cm}, clip, width=1\columnwidth]{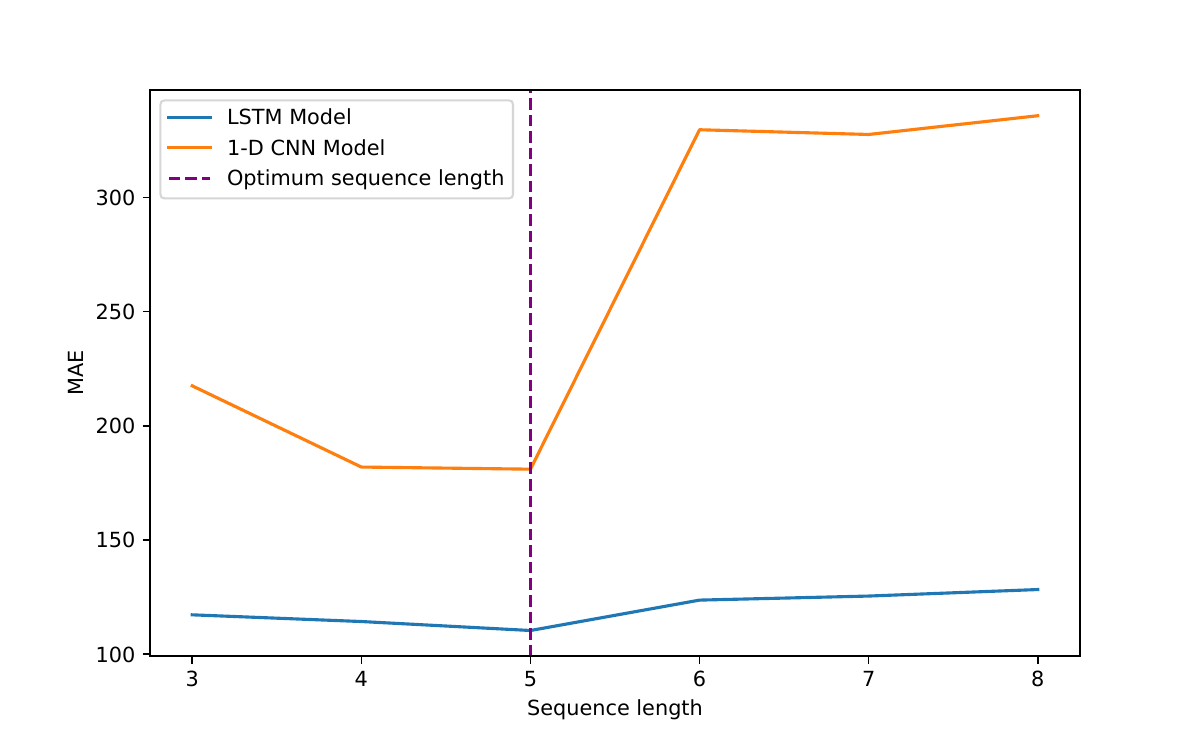}
 \caption{Analysing the impact of input sequence length in forecasting the oil production with respect to evaluation metric (MAE).}
 \vspace{-0.5cm}
 \label{figure:sequence}
\end{figure}

In time-series analysis, selecting the optimal sequence length highly influences the prediction's precision~\cite{9760052, 8671459, 9735273}. It requires a strong domain knowledge to comprehend the impact of previous data points in the data sequence. 
In this work, for the proposed 1-D CNN-based and LSTM-based oil production forecasting models, 
various sequence lengths (ranging from three to eight) are considered in selecting the optimal sequence length. 
Fig.~\ref{figure:sequence} shows the influence of sequence length in the proposed models' performance in terms of MAE. 
From these analyses, it is evident that when the sequence length is increased from three to five, the error in the forecast gradually decreases, but beyond that range, the error increases. As a result, a sequence length of five is chosen as the optimal value.

\subsubsection{Optimizing layer configuration}

To finalize an optimal model wrt the number of hidden layers and neurons several sanity analyses are conducted. For example, Table~\ref{table:hyper_tuning} summarizes the performances of ten different architectural configurations that led us to finalize the optimal model elaborated in Section~\ref{lstm-regressor} and Table~\ref{table:lstm_char}. 
It is clear from this table, the best performant models (i.e., with the lowest MAE and the highest R-squared) are model 3 and model 4, respectively for LSTM-based and 1-D-based regressors. 

For more information about the optimized LSTM network, Table~\ref{table:lstm_char} was provided to summarize the
model’s connectivity pattern with respective layer details. Furthermore, the best performance for 1-D CNN network was gained by model number 4, which includes three convolutions, one max pooling, one flatten and one dense layer.  A table was provided to conclude the details of the proposed CNN-based model.

\subsection{Model complexity analysis}
In deep learning, the process of passing a single input through a model to produce an output is called an inference. It is important to know the inference time of a model in advance because it allows researchers to design and optimize the model for better performance. To measure the inference time, the total number of computations performed by the model must be calculated. One way to do this is by using a measure called Floating Point Operations (FLOPs), which counts the number of operations involving floating point values in the model. The inference time can then be calculated by dividing the number of FLOPs by the number of FLOPs that the CPU can perform in a given amount of time.

\section{Overall Analysis}\label{sec-Overall analysis}

\subsection{Research Gap In the Existing Works}

In the existing works, there is a lack of experiments conducted on the Volve oil field production datasets. We believe it is because the dataset became public quite recently in 2018. In addition, all the existing works focus on well-specific model developments. Such models are not applicable to forecast the oil production in other wells. Hence, the existing works do not report their performances in all relevant evaluation metrics (MAE and R$^2$ score). However, this work develops a generalized model, which is applicable to all the oil production wells in the Volve oil field and the performance of the models is evaluated using MAE an R$^2$ score.



\subsection{Quantitative Analysis}

A thorough comparative analysis is conducted in comparison to the existing works, in Table~\ref{table:comparison}, where the standard linear regression described in Section~\ref{sec:baseline} is considered as the baseline. When compared to this baseline, the best existing work, the conventional neural network-based solution proposed by  Chahar~\textit{et al.}~\cite{chahar2022data} achieves $8.5\%$ improvement, while the proposed LSTM-based and 1-D CNN-based models provide significant improvement of $37\%$ and $14\%$, respectively. 
In addition to its superior performance, the complexity of the proposed LSTM-based regressor is $\simeq45\%$ less than that of the CNN-based counterpart in terms of the number of trainable parameters (cf. Table~\ref{table:cnn_char}, Table~\ref{table:lstm_char} and Table~\ref{table:hyper_tuning}). Hence, it requires only about $2\%$ of computations compared to the CNN-based counterpart in terms of FLOPs. Therefore, the proposed LSTM-based model is a more resource-friendly and efficient solution. The holistic analysis suggests that the LSTM-based oil production forecasting model is a robust, generalized, and reliable solution.
As it can be seen from Table~\ref{table:hyper_tuning}, LSTM models 2 and 3 have similar MAE and FLOPs. Model 2 is preferable for the resource-limited computational platform, while model 3 is a better solution when prediction precision is very important for the optimal operation of oil production forecasting.

\setlength{\tabcolsep}{2pt}

\begin{table*}[!ht]

\centering
\caption{A Quantitative Analysis. Comparing the Performances of the Proposed Models and the Existing Works. Note: In terms of comparison, (1) FLOPS and MAE, the Lower the Better, (2) R$^2$, the Higher the Better, (3) "\textemdash" Means No Information Is Found in the Respective Work, and (3) $\upuparrows$ - Average Percentage of Improvement Compared to the Baseline Model.}
\label{table:comparison}

\begin{tabular}{|c|c|c|c|c|c|c|c|c|c|}
\hline
\textbf{Network}  & \textbf{Error Metric} & \textbf{15/9-F-12 H} & \textbf{15/9-F-14 H} & \textbf{15/9-F-1 C} & \textbf{15/9-F-11 H} & \textbf{15/9-F-15 D} & \textbf{Global Test Set} & \textbf{FLOPs}  & \textbf{$\upuparrows$} \\ \hline
\multirow{2}{*}{Proposed LSTM Model} & \textbf{MAE} & 109.42& 73.77  & 177.64    & 133.49   & 156.74   & 111.16  & \multirow{2}{*}{33312} & \multicolumn{1}{c|}{\multirow{2}{*}{+37\%}} \\ 
& \textbf{$R^{2}$}  & 0.98  & 0.99  & 0.95  & 0.97  & 0.95  & 0.98   &  &\multicolumn{1}{c|}{}  \\ \hline
\multirow{2}{*}{Proposed 1-D CNN Model}  & \textbf{MAE}  & 157.04   & 106.62    & 201.29 & 186.58    & 196.52    & 151.64  & \multirow{2}{*}{3008544} & \multicolumn{1}{c|}{\multirow{2}{*}{+14\%}}\\
& \textbf{$R^{2}$}      & 0.97              & 0.98           & 0.95             & 0.94 & 0.91              & 0.96 &    &\multicolumn{1}{c|}{}                    \\ \hline
\multirow{2}{*}{Linear Regression Model}               & \textbf{MAE}          & 209.65          & 218.76          & 256.10         & 196.01          & 270.40          & 176.36            & \multirow{2}{*}{Not defined}  & \multirow{2}{*}{Base line}         \\
& \textbf{$R^{2}$}      & 0.94          & 0.91          & 0.79         & 0.90          & 0.90          & 0.92           &   &\multicolumn{1}{c|}{}          \\ \hline
 \multirow{2}{*}{{Neural Network~\cite{chahar2022data}}}   & \textbf{MAE}          & 161.24               & \multirow{2}{*}{\textemdash} & \multirow{2}{*}{\textemdash} & \multirow{2}{*}{\textemdash}  & \multirow{2}{*}{\textemdash} & \multirow{2}{*}{\textemdash} & \multirow{2}{*}{\textemdash}  & \multicolumn{1}{c|}{\multirow{2}{*}{+8.5\%}}   \\
& \textbf{$R^{2}$}      & 0.94                 &           &          &          &           &       &      &\multicolumn{1}{c|}{}       \\ \hline    
\multirow{2}{*}{{Random Forest~\cite{wang2021ensemble}}}    & \textbf{MAE}          & \textemdash          & \multirow{2}{*}{\textemdash} & \multirow{2}{*}{\textemdash} & \multirow{2}{*}{\textemdash}  & \multirow{2}{*}{\textemdash} & \multirow{2}{*}{\textemdash} & \multirow{2}{*}{\textemdash}  & \multicolumn{1}{c|}{\multirow{2}{*}{-1\%}}  \\
 & \textbf{$R^{2}$}      & 0.91  &   &   &   &   &   &   &\multicolumn{1}{c|}{}          \\ \hline
\multirow{2}{*}{{Decline Curve~\cite{masini2019decline}}} & \textbf{MAE}  &  \multirow{2}{*}{\textemdash}  & \textemdash   & \multirow{2}{*}{\textemdash} & \multirow{2}{*}{\textemdash}  & \multirow{2}{*}{\textemdash} & \multirow{2}{*}{\textemdash} & \multirow{2}{*}{\textemdash} &\multicolumn{1}{c|}{\multirow{2}{*}{-10\%}} \\
& \textbf{$R^{2}$}  &   & 0.82 &  &  &  &    &     &\multicolumn{1}{c|}{}        \\ \hline
\multirow{2}{*}{{AdaBosst~\cite{noshi2019intelligent}}}  & \textbf{MAE}  & \multirow{2}{*}{\textemdash} & 300 & \multirow{2}{*}{\textemdash} & \multirow{2}{*}{\textemdash}  & \multirow{2}{*}{\textemdash} & \multirow{2}{*}{\textemdash} & \multirow{2}{*}{\textemdash}  &\multicolumn{1}{c|}{\multirow{2}{*}{-70\%}} \\
 & \textbf{$R^{2}$} &  & \textemdash          &  &  &  &   &     &\multicolumn{1}{c|}{}        \\ \hline
\multirow{2}{*}{{LSTM~\cite{ng2022well}}}  & \textbf{MAE} & \multirow{2}{*}{\textemdash}  & \textemdash  & \multirow{2}{*}{\textemdash} & \multirow{2}{*}{\textemdash}  & \multirow{2}{*}{\textemdash} & \multirow{2}{*}{\textemdash} & \multirow{2}{*}{\textemdash} & \multicolumn{1}{c|}{\multirow{2}{*}{+5\%}}        \\
& \textbf{$R^{2}$}      &         & 0.97 &  &  &  &   &      &\multicolumn{1}{c|}{}       \\ \hline

\end{tabular}
\end{table*}

\subsection{Qualitative Analysis}

\begin{figure*}[!t]
 \centering
 \includegraphics[width=1\textwidth]{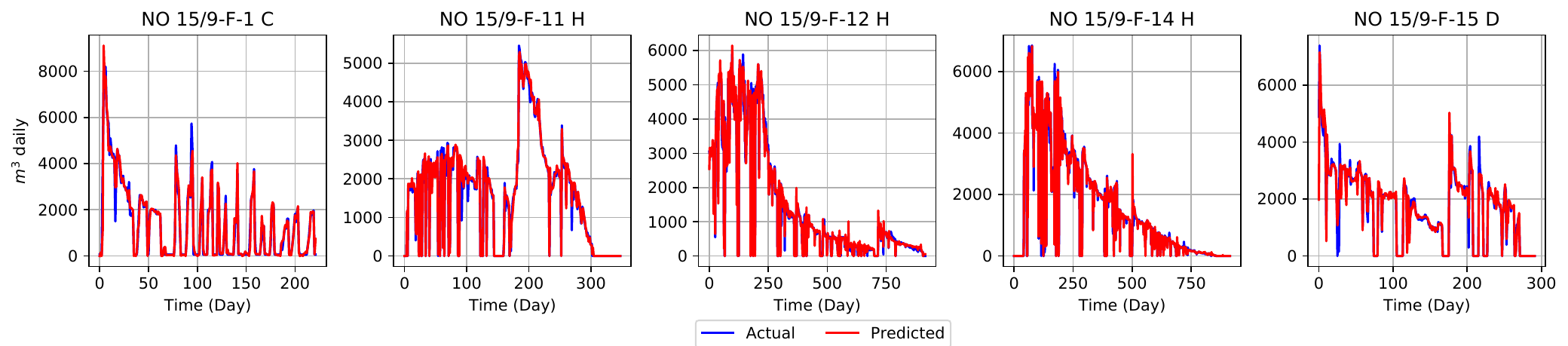}
 \caption{Qualitative analysis I. Plots of the proposed LSTM-based oil production forecasting along with the ground truth values for all five production wells, No 15/9-F-1 C, No 15/9-F-11 H, No 15/9-F-12 H, No 15/9-F-14 H, and No 15/9-F-15 D, in the Volve oil field.}
 \label{figure:lstmnet}

\end{figure*}

\begin{figure*}[!t]
 \centering
 \includegraphics[width=1\textwidth]{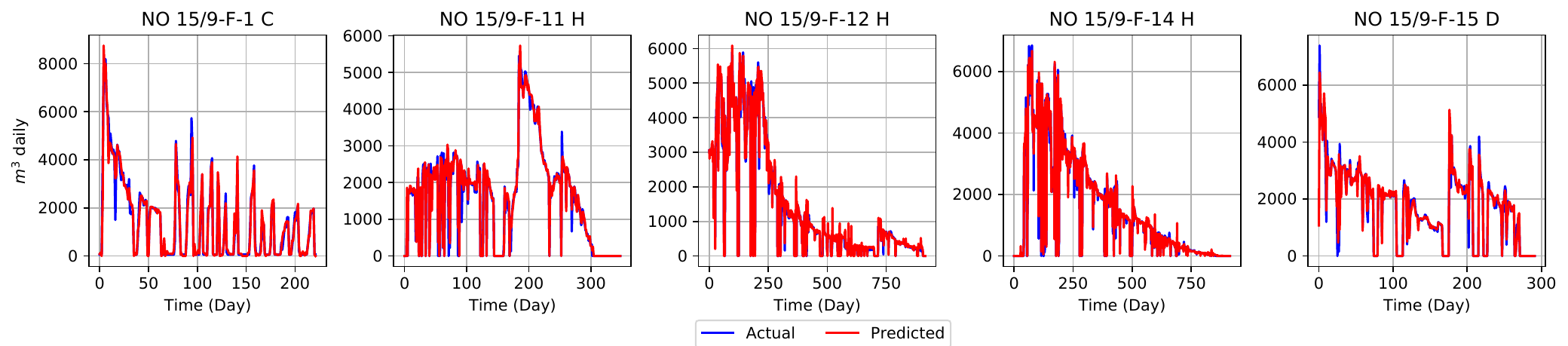}
 \caption{Qualitative analysis II. Plots of the proposed 1D CNN-based oil production forecasting along with the ground truth values for all five production wells, No 15/9-F-1 C, No 15/9-F-11 H, No 15/9-F-12 H, No 15/9-F-14 H, and No 15/9-F-15 D, in the Volve oil field.}
 \label{figure:cnnnet}
\end{figure*}

Figures~\ref{figure:lstmnet} and \ref{figure:cnnnet} compare the proposed models' oil production forecasting results with the ground truths of the respective test sequences wrt the five production wells in the Volve oil field. From the plots of the prediction and actual values, one can observe that the proposed models' forecasts are very close to the actual values. It is further verified by the quantitative comparisons given in Table~\ref{table:comparison}. 

\section{Conclusion}\label{sec-conclusion}

In this study, an LSTM-based model and a 1-D CNN-based model were proposed for the time series production forecasting of the Volve oilfield. To determine the appropriate sequence length in the time series data analysis, a comprehensive investigation was conducted at the outset. After the best model topology was chosen based on the hyper-parameter tuning procedure, it was found that the LSTM-based model had better performance than the 1-D CNN model, as demonstrated by the MAE, R-squared, and complexity metric. From an applied perspective, since data from all of the wells was used in the training and testing of the models, they can be generalized to the other existing wells. However, it is important to note that the generalizability of the models has not been investigated in this paper and is a topic for future research. In addition, another potential direction for future research in oil production forecasting using deep neural networks could be the integration of additional data sources. For example, incorporating data on well activity, drilling plans, and geological information could potentially improve the accuracy of forecasts.

\bibliographystyle{ieeetr}
\bibliography{references}

\end{document}